\documentclass[11pt,a4paper]{article}
\usepackage[hyperref]{acl2020}

\usepackage{microtype}

\usepackage{times}
\usepackage{latexsym}
\usepackage{booktabs}
\usepackage{xspace}
\usepackage{xcolor}
\usepackage{soul}
\usepackage{algorithm}
\usepackage{algorithmicx}
\usepackage{amsmath}
\usepackage[frozencache]{minted}
\usepackage{bm}
\usepackage{tikz}
\usepackage{inconsolata}
\usepackage[T1]{fontenc}
\usepackage{gb4e}
\usepackage{array}
\usepackage{ragged2e}
\usepackage{tabularx}

\usepackage{url}

\aclfinalcopy % Uncomment this line for the final submission
\def\aclpaperid{54}

\newcommand\thiswork{\textsc{geca}\xspace}

\newcommand\scan{\textsc{scan}\xspace}
\newcommand\nacs{\textsc{nacs}\xspace}
\newcommand\geoquery{\textsc{GeoQuery}\xspace}

\newcommand\translate{$\triangleright$\xspace}

\newcommand\sigA{\textsuperscript{$\dagger$}}
\newcommand\sigB{\textsuperscript{$\ddagger$}}
\newcommand\nosig{\phantom{\textsuperscript{$\dagger$}}}
\newcommand\spm[1]{{\small $\pm$ #1}}

\newcommand\eref[1]{(\ref{#1})}
\newcommand\hlt[1]{\underline{\smash{#1}}}

\title{Good-Enough Compositional Data Augmentation}

\author{Jacob Andreas \\
  MIT CSAIL \\
  {\tt jda@mit.edu}
}

\date{}

\begin{document}
\maketitle
\begin{abstract}
  We propose a simple data augmentation protocol aimed at providing a
  compositional inductive bias in conditional and unconditional sequence models.
  Under this protocol, synthetic training examples are constructed by taking
  real training examples and replacing (possibly discontinuous) fragments with
  other fragments that appear in at least one similar environment. The protocol
  is model-agnostic and useful for a variety of tasks. Applied to neural
  sequence-to-sequence models, it reduces error rate by as much as 87\% on 
  diagnostic tasks from the \scan dataset and 16\% on a semantic parsing task.
  Applied to n-gram language models, it reduces perplexity by roughly 1\% on
  small corpora in several languages.
\end{abstract}

\section{Introduction}

This paper proposes a rule-based data augmentation protocol for sequence
modeling. Our approach aims to supply a simple and model-agnostic bias
toward compositional reuse of previously observed sequence fragments in novel
environments. Consider a language modeling task in which we wish to estimate a probability 
distribution over a family of sentences with the following finite sample as 
training data:
\begin{exe}
	\ex \label{ex:gen-train} \begin{xlist}
		\ex {\it The cat sang.}
		\ex {\it The wug sang.}
		\ex {\it The cat daxed.}
	\end{xlist}
\end{exe}
In language processing problems, we often want models to generalize beyond this
dataset and infer that \eref{ex:gen-test-good} is also probable but
\eref{ex:gen-test-bad} is not:
\begin{exe}
  \ex \label{ex:gen-test} \begin{xlist} 
    \ex \label{ex:gen-test-good} {\it The wug daxed.}
    \ex \label{ex:gen-test-bad} {\it * The sang daxed.}
  \end{xlist}
\end{exe}
This generalization amounts to an inference about syntactic categories
\citep{Clark00Categories}. Because \emph{cat} and \emph{wug} are interchangeable
in (\ref{ex:gen-train}a) and (\ref{ex:gen-train}b), they are also likely
interchangeable elsewhere; \emph{cat} and \emph{sang} are not similarly
interchangeable. Human learners make judgments like
\eref{ex:gen-test} about novel lexical items \citep{Berko58Wug} and fragments
of novel languages \citep{Lake19HumanComp}. But we do not
expect such judgments from unstructured generative models trained to maximize the
likelihood of the training data in \eref{ex:gen-train}.

A large body of work in natural language processing provides generalization to
data like \eref{ex:gen-test-good} by adding structure to the learned predictor
\citep{Chelba98SynLm,Chiang05Hierarchical,Dyer16RNNG}.
On real-world datasets, however, such models are typically worse than
``black-box'' function approximators like neural networks, even for black-box
models that fail to place probability mass on either example in
\eref{ex:gen-test} given small training sets like \eref{ex:gen-train}
\citep{Melis18LMEval}.
To the extent that we believe \eref{ex:gen-test-good} to capture an important
inductive bias, we would like to find a way of softly encouraging it without
tampering with the structure of predictors that work well at scale.
In this paper, we introduce a procedure for
generating synthetic training examples by recombining real ones, such that
\eref{ex:gen-test-good} is assigned non-negligible probability because it
\emph{already appears in the training dataset}.

\begin{figure}[b]
  \vspace{-1.5em}
  \center
  \includegraphics[width=\columnwidth,clip,trim=0.1in 5.5in 4in 0in]{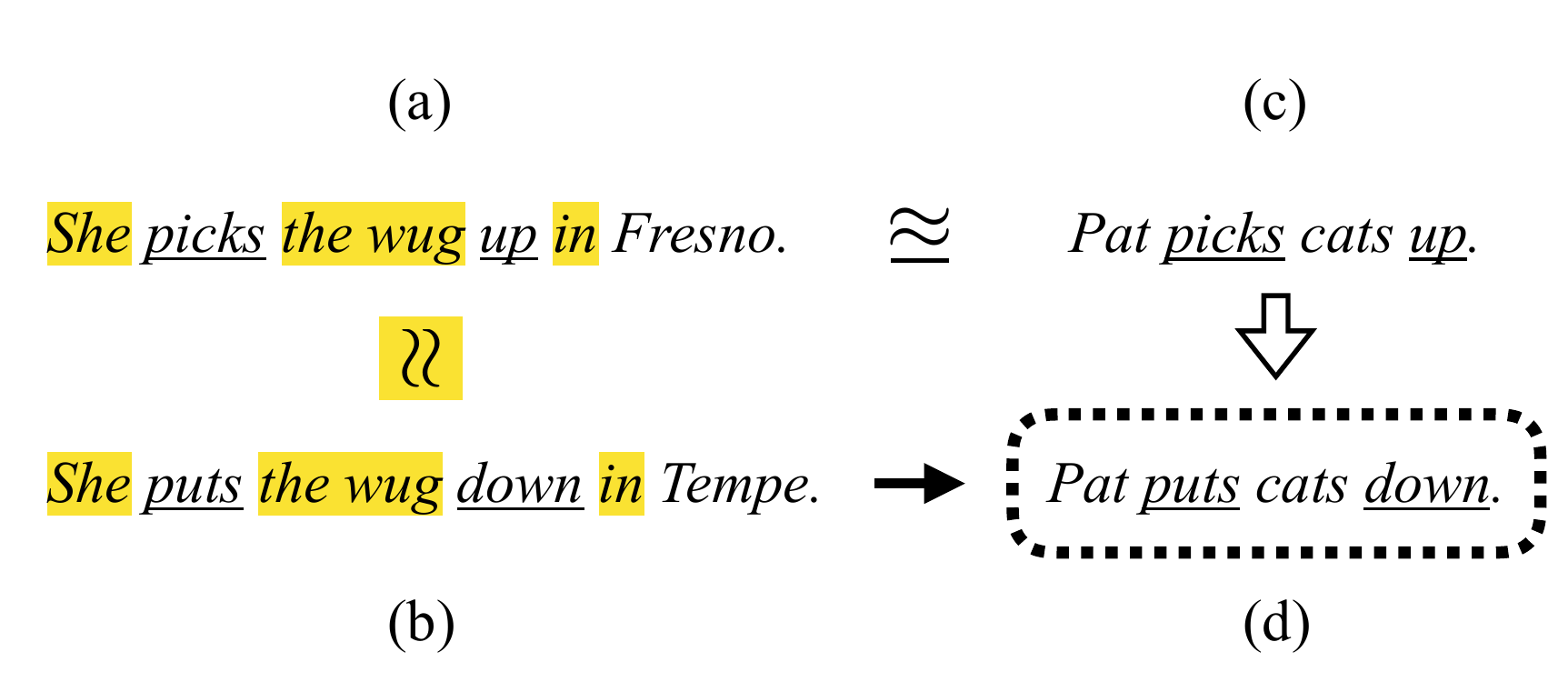}
  \vspace{-2.0em}
  \caption{
    Visualization of the proposed approach: two discontinuous sentence fragments
    (a--b, underlined) which appear in similar environments (a--b, highlighted)
    are identified.  Additional sentences in which the first fragment appears
    (c) are used to synthesize new examples (d) by substituting in the second
    fragment.
  }
	\label{fig:teaser}
  \vspace{-4pt}
\end{figure}

The basic operation underlying our proposal (which we call \thiswork, for
``good-enough compositional augmentation'') is depicted in
\autoref{fig:teaser}: if two (possibly discontinuous) fragments of training
examples appear in \emph{some} common environment, then \emph{any} additional
environment where the first fragment appears is also a valid environment for
the second. 

\thiswork is crude: as a linguistic principle, it is both limited and imprecise.
As discussed in Sections \ref{sec:ling} and \ref{sec:scan}, it captures a narrow
slice of the many phenomena studied under the heading of ``compositionality'',
while also making a number of incorrect predictions about real language data.
Nevertheless, \thiswork appears to be quite effective across a range of learning
problems. In semantic parsing, it gives improvements comparable to the
task-specific data
augmentation approach of \citet{Jia16Recombination} on logical
expressions, better performance than that approach on a different split of the
data designed to test generalization more rigorously, and corresponding improvements
on a version of the dataset with a different meaning
representation language. Outside of semantic parsing, it solves two
representative problems from the \scan dataset of \citet{Lake18SCAN} that are
synthetic but precise in the notion of compositionality they test. Finally, it
helps with some (unconditional) low-resource language modeling problems in a
typologically diverse set of six languages.

\section{Background}
\label{sec:background}

Recent years have seen tremendous success at natural language transduction and
generation tasks using complex function approximators, especially recurrent
\citep{Sutskever14NeuralSeq} and attentional \citep{Vaswani17Transformer} neural
models. With enough training data, these models are often more accurate than
than approaches built on traditional tools like regular transducers and
context-free grammars \citep{Knight05Transducers}, which are brittle and
difficult to efficiently infer from large datasets.

However, models equipped with an explicit symbolic generative process have at
least one significant advantage over the aforementioned black-box approaches:
given a grammar, it is straightforward to precisely characterize how that
grammar will extrapolate beyond the examples in a given training set to
out-of-distribution data. Indeed, it is often possible for researchers to
design the form that this extrapolation will take: smoothed n-gram language
models ensure that no memorization is possible beyond a certain length
\citep{Ney94Smoothing}; CCG-based semantic parsers can make immediate use of
entity lexicons without having ever seen the lexicon entries used in real
sentences \citep{Zettlemoyer05CCG}.

It is not the case
that
black-box neural models are fundamentally incapable of this kind of
predictable generalization---the success of these models at capturing
long-range structure in text \citep{Radford19GPT2} and controlled algorithmic
data \citep{Graves14NTM} indicate that  some representation of hierarchical
structure can be learned given enough data.  But the precise point at which
this transition occurs is not well characterized, and it is evidently beyond the
scale available in many real-world problems.

How can we improve the behavior of high-quality black-box models in these
settings? There are many sophisticated tools available for improving the
function approximators or loss functions themselves---structured regularization
of parameters
\citep{Oh17ZeroShot}, posterior regularization \citep{Ganchev10PR,Hu18DeepPR},
explicit stacks \citep{Gref15Stack} and composition operators
\citep{Bowman16SPINN,Russin19SynAtt}. These existing proposals tend to be task-
and architecture-specific.  But to the extent that the generalization problem
can be addressed by increasing the scale of the training data, it is natural to
ask whether we can address the problem by increasing this scale
\emph{artificially}---in other words, via data augmentation. 

Data augmentation techniques, which generate auxiliary training data by
performing structured transformation or combination of training examples, are
widely used in computer vision \cite{Krishevsky12ImageNet,
Zhang17Mixup,Summers19Aug}. Within NLP, several data augmentation approaches
have been proposed for text classification 
\citep[e.g.][]{Ratner17TF,Wei19Aug}; these approaches give improvements
even when combined with large-scale pretraining \citep{Hu19Aug}.
\citet{Jia16Recombination} study data augmentation and compositionality in
specific setting of learning language-to-logical-form mappings, beginning from
the principle that data is compositional if it is generated by an explicit
grammar that relates strings to logical forms. This view of compositionality as
determined by synchronicity between form and meaning is essentially Montagovian
and well-suited to problems in formal semantics \cite{Montague73PTQ}; however,
it requires access to structured meaning representations with explicit types
and bracketings, which are not available in most NLP applications. 

Here we aim at a notion of compositionality that is simpler and more general: a
bias toward identifying recurring fragments seen at training time, and re-using
them in environments distinct from those in which they were first observed.
This view makes no assumptions about the availability of brackets and types,
and is synchronous only to the extent that the notion of a fragment is
permitted to include content from both the source and target sides.  We will
find that it is nearly as effective as existing approaches in the specific
settings for which they were designed, but also effective on a variety of
problems where they cannot be applied.

\section{Approach}
\label{sec:approach}

Consider again the example in \autoref{fig:teaser}. Our
data augmentation protocol aims to discover substitutable sentence
\textbf{fragments} (underlined), with the fact that a pair of fragments appear in
some common sub-sentential \textbf{environment} (highlighted) taken as evidence
that the fragments belong to a common category.  To generate a new examples for
the model, an occurrence of one fragment is removed from a sentence to produce
a sentence \textbf{template}, which is then populated with the other
fragment.

Why should we expect this procedure to produce well-formed training examples?
The existence of syntactic categories, and the expressibility of
well-formedness rules in terms of these abstract categories, is one of the
foundational principles of generative approaches to syntax
\citep{Chomsky14Aspects}. The observation that context provides a
strong signal about a sentence fragment's category is in turn the foundation of
distributional techniques for the study of language \citep{Firth57Synopsis}.
Combining the two gives the outlines of the above procedure.

This combination has a productive history in natural language processing: when
fragments are single words, it yields class-based language models
\citep{Brown92LM}; when fragments are contiguous spans it yields
unsupervised parsers \citep{Clark00Categories,Klein02CCM}. The present data
augmentation scenario is distinguished mainly by the fact that we are
\emph{unconcerned} with producing a complete generative model of data, or with
recovering the latent structure implied by the presence of nested syntactic
categories. We can still synthesize high-precision examples of well-formed
sequences by identifying individual substitutions that are likely to be correct
without understanding how they fit into the grammar as a whole.

Indeed, if we are not concerned with recovering linguistically plausible
analyses, we need not limit ourselves to words or contiguous sentence
fragments. We can take
\begin{exe}\ex\begin{xlist}
  \ex {\it She picks the wug up.}
  \ex {\it She puts the wug down.}
\end{xlist}\end{exe}
as evidence that we can use \textit{picks\ldots{}up} wherever we can use
\textit{puts\ldots{}down}. Indeed, given a \emph{translation} dataset:
\begin{exe}\ex\label{ex:translation}\begin{xlist}
  \ex {\it I sing.\ \translate Canto.}
  \ex {\it I sing marvelously.\ \translate \\ Canto maravillosamente.}
  \ex {\it I dax marvelously.\ \translate \\ Dajo maravillosamente.}
\end{xlist}\end{exe}
we can apply the same principle to synthesize \textit{I dax.\ \translate
Dajo.} based on the common environment
\textit{\ldots{}marvelously \translate \ldots{}maravillosamente}.  From the perspective of a
generalized substitution principle, the alignment problem in machine translation
is the same as the class induction problem in language modeling, but with
sequences featuring large numbers of gappy fragments and a boundary symbol
\translate.

The only remaining question is what makes two environments similar enough to
infer the existence of a common category. There is, again, a large literature on
this question (including the aforementioned work in language modeling,
unsupervised parsing, and alignment), but in the current work we will make use
of a very simple criterion: 
fragments are interchangeable if they occur in at least one \textbf{lexical
environment} that is exactly the same. 

Given a \textbf{window size} $k$ and sequence of $n$ tokens $w = w_1 w_2 \cdots
w_n$, define a \textbf{fragment} as a set of non-overlapping spans of $w$, a
\textbf{template} as a version of $w$ with a fragment removed, and an
\textbf{environment} as a template restricted to a $k$-word window around each
removed fragment. Formally, (letting $[i, j]$ denote $\{i, i+1, \ldots, j\}$) we have:
\begin{align}
  & \hspace{-.5em}\texttt{fragments}(w) = \{ \{ w_{a_1..b_1}, w_{a_2..b_2}, \ldots \}: \nonumber \\
  & ~~~1 \leq a_i < b_i \leq n \textrm{, all } [a_i, b_i] \textrm{ disjoint} \}
  \\[.5em]
  & \hspace{-.5em}\texttt{tpl}(w, f) = ( w_j:
  \forall w_{a_i..b_i} \in f. ~ j \not\in [a_i, b_i] ) \\[.5em]
  & \hspace{-.5em}\texttt{env}(w, f) = \{ w_j: \nonumber \\
  & ~~~w_j \in \texttt{tpl}(w, f) \textrm{ and } \nonumber \\
  & ~~~\exists w_{a_i..b_i} \in f. ~ j \in [a_i - k, b_i + k] \}
\end{align}

In \autoref{fig:teaser}(a), the underlined
\emph{\underline{picks}\ldots\underline{up}} is one possible fragment that could
be extracted from the sentence. The corresponding template is \emph{She\ldots
the wug \ldots in Fresno}, and with $k=1$ the environment is \emph{She\ldots the
wug \ldots in}.
As shown in \autoref{fig:teaser}(d), any fragment may be substituted into any
template with the same number of holes. Denote this substitution operation by
$t / f$.
The data augmentation operation that defines \thiswork is formally
stated as follows:
\begin{center}
  \vspace{-1em}
\noindent \fbox{\parbox{0.98\columnwidth}{
  If the training data contains sequences $w = t_1/f_1$, $x = t_1'/f_1$ and
  $y = t_2/f_2$, with $env(w, t_1) = env(y, t_2)$ and $t_1' \neq t_1$, then
  synthesize a new training example $z = t_1'/f_2$.
}}
\end{center} \strut\\[-1em]
If a fragment occurs multiple times within a given example, all instances are
replaced (see \autoref{fig:scan-synth}).

% SCAN
\begin{table*}
  \center
\begin{tabular}{lcccc}
\toprule
& jump / \scan & jump / \nacs & right / \scan & right / \nacs \\
\midrule
  seq2seq & 0.00 \spm{0.00} & 0.00 \spm{0.00} & 0.00 \spm{0.00} & 0.00 \spm{0.00} \\
  + \thiswork & {\bf 0.87} \spm{0.02} & {\bf 0.67} \spm{0.01} & {\bf 0.82} \spm{0.04} & {\bf 0.82} \spm{0.03}
\\
\bottomrule
\end{tabular}
  \caption{
    Sequence match accuracies on \scan
    datasets, in which the learner must generalize to new compositional uses of a
    single lexical item (``jump'') or multi-word modifier (``around right'') when
    mapping instructions to action sequences (\scan) or vice-versa
    \citep[\nacs,][]{Bastings18NACS}.  While the sequence-to-sequence model is
    unable to make any correct generalizations at all, applying \thiswork enables
    it to succeed most of the time. Scores are averaged over 10 random seeds;
    the standard deviation across seeds is shown. All improvements are
    significant (paired binomial test, $p \ll 0.001$).
  }
\vspace{-1em}
\label{tab:scan}
\end{table*}

\paragraph{Linguistic notes}
\label{sec:ling}
Despite the fact that the above operation is motivated by insights from
generative syntax and distributional semantics, it should be emphasized that it
is, as a statement of a general linguistic principle, obviously wrong.
Counterexamples abound: in English, stress-derived nouns (e.g.\
\textit{r\'ecord} from \textit{rec\'ord}) will be taken as evidence that many
nouns and verbs are interchangeable; in Mandarin Chinese, \emph{k\v{e}sh\`i} and
\emph{d\`ansh\`i} both mean ``but'', but \emph{k\v{e}sh\`i} alone can be
used in particular constructions to mean ``very''.

What ultimately matters is the relative frequency of such errors: if their
contribution to an inaccurate model is less than the inaccuracy caused by the
original shortage of training data, the \thiswork still helps. In
conditional problems, like the machine translation example above, such errors
may be totally harmless: if we synthesize a new $(x, y)$ pair with $x$ outside
the support of the real training data, they may not influence the model's
predictions on the true support beyond providing useful general inductive bias.

\paragraph{Implementation}
\label{sec:impl}

Na\"ive implementation of the boxed operation takes $O(t^3 f^3)$ time (where $t$
is the number of distinct templates in the dataset and $f$ the number of
distinct fragments). This can be improved to $O(ft^2e)$ (where
$e$ is the number of templates that map to the same environment) by building
appropriate data structures (\autoref{algo}).

\begin{algorithm}[b!]
\footnotesize
\vspace{1em}
\begin{minted}{python}
f2t = dict(default=set()) # fragment -> template
t2f = dict(default=set()) # template -> fragment
e2t = dict(default=set()) # env -> template
for seq in dataset:
  for f in fragments(seq): # Eq. 1
    template = tpl(seq, fragment) # Eq. 2
    add(f2t[fragment], template)
    add(t2f[template], fragment)
    add(e2t[env(seq, fragment)], template) # Eq. 3

t2t = dict(default=set())
for fragment in keys(f2t)):
  for template in f2t[fragment]:
    for template2 in f2t[fragment]:
      for new_template in e2t[env(template2)]
        add(t2t[template1], new_template)

for template1, template2 in t2t:
  for fragment in t2f[template1]
    if fragment not in t2f[template2]:
      yield template2 / fragment
\end{minted}
\caption{Sample \thiswork implementation.}
\label{algo}
\end{algorithm}

Space requirements might still be considerable (comparable to those used by
n-gram language models), and strategies from the language modeling literature
can be used to reduce memory usage \cite{Heafield11KenLM}. This algorithm is
agnostic with respect to the choice of fragmentation and environment functions;
task-specific choices are described in more detail for each experiment below.

\section{Diagnostic experiments}
\label{sec:scan}

\begin{figure}[b!]
  \center
  \footnotesize
  \begin{tabularx}{\columnwidth}{X}
    \toprule
    \it walk             \\ \tt WALK \\[.5em]
    \it walk left twice  \\ \tt LTURN WALK LTURN WALK \\[.5em]
    \it jump             \\ \tt JUMP \\[.5em]
    \it jump around left \\ \tt LTURN JUMP LTURN JUMP LTURN JUMP LTURN JUMP \\[.5em]
    \it walk right       \\ \tt RTURN WALK \\
    \bottomrule
  \end{tabularx}
  \caption{Example \scan data. Each example consists of a synthetic natural
  language command (left) paired with a discrete action sequence (right).}
  \label{fig:scan-example}
\end{figure}

We begin with a set of experiments on synthetic data designed to precisely test
whether \thiswork provides the kind of generalization it was designed for. Here
we use the \scan dataset \cite{Lake18SCAN}, which consists of simple English
commands 
paired with sequences of discrete actions (\autoref{fig:scan-example}).
We focus
specifically on the \emph{add primitive (jump)} and \emph{add template (around
right)} conditions, which test whether the agent can be exposed to individual
commands or modifiers (e.g.\ \textit{jump} \translate \texttt{JUMP}) in isolation at
training time, and incorporate them into more complex commands like the earlier
example at test time.

We extract fragments with one gap and a maximum length of 4 tokens. The
environment is taken to be the complete template. Generated examples are
appended to the original dataset. As an example of the effect of this
augmentation procedure, the original \emph{jump} split has 12620 training
examples; \thiswork generates an additional 395 using 395 distinct templates and
6 distinct fragments.

With the original and augmented datasets, we train a one-layer LSTM
encoder--decoder model with an embedding size of 64, a hidden size of 512, a
bidirectional encoder and an attentional decoder
\citep{Hochreiter97LSTM,Bahdanau14Attention}. The model is trained using
\textsc{Adam} \citep{Kingma14Adam} with a step size of $0.001$ and a dropout
rate of $0.5$.

Results are shown in \autoref{tab:scan}.  In line with the original experiments
of \citeauthor{Lake18SCAN}, the baseline sequence-to-sequence model completely
fails to generalize to the test set. Applying \thiswork allows the learned model
to successfully make most tested generalizations across single and multi-word
entries, and in both instruction-to-action and action-to-instruction directions.

\paragraph{Analysis: examples}

Some synthesized examples are shown in \autoref{fig:scan-synth}. Success at the
\emph{add primitive} condition stems from the constraint that the single example
usage of the primitive must still be a valid (command, action) pair, and all
verbs are valid commands in isolation. Only three examples---\textit{run} \translate
\texttt{RUN}, \textit{walk} \translate \texttt{WALK} and \textit{look} \translate
\texttt{LOOK}---provide the evidence that \thiswork uses to synthesize to new
usages of \textit{jump}; if these were removed, the sequence-to-sequence model's
training accuracy would be unchanged but \thiswork would fail to synthesize any
new examples involving \textit{jump}, and test accuracy would fall to zero.
For the \emph{add template} condition, \thiswork correctly replaces all
occurrences of \texttt{LTURN} with \texttt{RTURN} to produce new examples of the
\emph{around right} template; this example highlights the usefulness of
\thiswork's ability to discover discontinuous and non-context-free substitutions.

\paragraph{Analysis: dataset statistics}

To further understand the behavior of \thiswork, we conduct a final set of
analyses quantifying the overlap between the synthesized data and the
held-out data. We first measure \textbf{full example overlap}, the fraction of
test examples that appear in the augmented training set. (By design, no overlap
exists between the test set and the original training set.) After applying
\thiswork, 5\% of test examples for the \emph{add primitive} condition and 1\%
of examples for the \emph{add template} condition are automatically synthesized.
Next we measure \textbf{token co-occurrence overlap}: we compute the set of
(input or output) tokens that occur together in any test example, and then
measure the fraction of these pairs that also occur together in some training
example. For the \emph{add primitive} condition, \thiswork increases token
co-occurrence overlap from 83\% to 96\%; for the \emph{add template} condition it
is 100\% even prior to augmentation.

\begin{figure}
  \center
  \footnotesize
  \begin{tabularx}{\columnwidth}{>{\RaggedRight\arraybackslash}X}
    \toprule
    {\bfseries\itshape add primitive (jump)} \\[0.5em]
    \it \hlt{walk} thrice after \hlt{walk} right \\
    \tt RTURN \hlt{WALK} \hlt{WALK} \hlt{WALK}
    \hlt{WALK} \\[.5em]
    \it jump opposite left thrice after turn opposite right  \\
    \tt RTURN RTURN LTURN LTURN \hlt{JUMP} LTURN
    LTURN \hlt{JUMP} LTURN LTURN \hlt{JUMP} \\
    \midrule
    {\bfseries\itshape add template (around right)} \\[0.5em]
    \it \hlt{look} right twice and turn opposite right twice \\
    \tt RTURN \hlt{LOOK} RTURN \hlt{LOOK} RTURN RTURN RTURN RTURN
    \\[0.5em]
    \it run around \hlt{right} and walk opposite \hlt{right} twice \\
    \tt \hlt{RTURN} RUN \hlt{RTURN} RUN \hlt{RTURN} RUN \hlt{RTURN} RUN
    \hlt{RTURN} \hlt{RTURN} WALK \hlt{RTURN} \hlt{RTURN} WALK \\
    \bottomrule
  \end{tabularx}
  \caption{Examples synthesized for the \scan tasks. Underlined words belong to
  the filled-in fragment; the remaining text is the template. \thiswork
  synthesizes some examples that exactly capture the desired generalization, and
  some examples that are unrelated.}
  \label{fig:scan-synth}
\end{figure}

It is important to note that \thiswork, which sees only the training set, is
unaware that some subset of the data is singled out for generalization
testing at evaluation time. The data augmentation protocol generates a large
number of spurious training examples unrelated to the desired
generalization (e.g.\ the first example in \autoref{fig:scan-synth}); however,
it also generates enough new usages of the target concept that the learner
generalizes successfully.

\begin{figure*}
  \center
  \footnotesize
  \scalebox{0.95}{
  \begin{tabularx}{1.03\textwidth}{>{\RaggedRight\arraybackslash}X}
    \toprule
    {\bfseries Logical forms} \\[0.5em]
    \it what is the lowest point in \hlt{rhode island} \\
    \tt ( A , lowest ( A , ( place ( A ) , loc ( A , B ) , const ( B , stateid (
    \hlt{rhode island} ) ) ) ) ) \\[0.5em]
    \it what states does the \hlt{florida} run through \\
    \tt ( A , ( state ( A ) , const ( B , riverid ( \hlt{florida} ) ) , traverse ( B , A ) ) ) \\[0.5em]
    \it what state borders the state with the \hlt{lowest population density} \\
    \tt ( A , ( state ( A ) , next\_to ( A , B ) , smallest ( C , ( state ( B )
    \hlt{, density} ( B , C ) ) ) ) ) \\
    \midrule
    {\bfseries SQL queries} \\[0.5em]
    \it what rivers run through west \hlt{wyoming} \\
    \tt SELECT RIVER0.NAME FROM RIVER AS RIVER0 WHERE RIVER0.TRAVERSE = " west
    \hlt{wyoming} " \\[0.5em]
    %\it what is the \underline{capital} of idaho \\
    %\tt \underline{SELECT STATE0.CAPITAL} FROM STATE AS STATE0 WHERE STATE0.STATE\_NAME = " idaho " ; \\
    \it which states have \hlt{towns} major \hlt{named} \hlt{springfield} \\
    \tt SELECT CITY0.STATE\_NAME FROM CITY AS CITY0 WHERE CITY0.NAME = "
    \hlt{springfield} " AND CITY0.POP > 150000 \\[0.5em]
    \it what is the population of the \hlt{area} of the largest state \\
    \tt SELECT CITY0.POP FROM CITY AS CITY0 WHERE
    CITY0.NAME = ( \hlt{SELECT STATE0.AREA} FROM STATE AS STATE0
    WHERE STATE0.AREA = ( SELECT MAX ( STATE1.AREA ) FROM STATE AS
    STATE1 ) ) \\
    \bottomrule
  \end{tabularx}
  }
  \caption{Examples synthesized for semantic parsing on \geoquery. Substituted
  fragments are underlined.
  \thiswork aligns named entities to their logical representations and abstracts
  over predicates.  Sometimes (as in the final example) synthesized examples are
  semantically questionable but have plausible hierarchical structure.}
  \label{fig:geo-synth}
\end{figure*}

\section{Semantic parsing}
\label{sec:semparse}

Next we turn to the problem of \emph{semantic parsing}, which has also been
a popular subject of study for questions about compositionality, generalization,
and data augmentation.  For the reasons discussed in \autoref{sec:ling}, we
expect qualitatively different behavior from this approach on real language data
without the controlled vocabulary of \scan. 

We study four versions of the \geoquery dataset \cite{Zelle95GeoQuery}, which
consists of 880 English questions about United States geography, paired with
meaning representations in the form of either logical expressions or
SQL queries. The standard train--test split for this dataset ensures that no
natural language \emph{question} is repeated between the train and test sets. As
\citet{Finegan18SQL} note, this provides only a limited test of generalization,
as many test examples feature a logical form that overlaps with the training
data; they introduce a more challenging \emph{query} split to ensure that
neither questions nor logical forms are repeated (even after anonymizing named
entities). 

We extract fragments with at most 2 gaps and at most 12 tokens. On the SQL query
split, the original training set contains 695 examples. \thiswork generates an
additional 1055 using 839 distinct templates and 379 distinct fragments.  For
the question split we use the baseline model of \citet{Jia16Recombination}; for
the query split we use the same sequence-to-sequence model as used for \scan and
introduce the supervised copy mechanism of \citet{Finegan18SQL}.  Environments
are again taken to be identical to templates. 

Results are shown in \autoref{tab:semparse}.
On the split for which \citet{Jia16Recombination} report results, \thiswork
achieves nearly the same improvements with weaker domain assumptions. On the
remaining splits it is more accurate.

\footnotetext[1]{In some cases these averages are slightly lower than the
single-run results previously reported in the literature. Note also that the
original publication from \citeauthor{Jia16Recombination} reports
\emph{denotation} accuracies; the results here are taken from their accompanying
code release. Overall trends across systems are comparable using either
evaluation metric.}

\paragraph{Analysis: examples}

Synthesized examples for the logical and SQL representations are shown in
\autoref{fig:geo-synth}. Despite the fact that the sequence-to-sequence model
uses neither gold entities or specialized entity linking machinery, the
augmentation procedure successfully aligns natural language entity names to
their logical representations and generalizes across entity choices. This
procedure also produces plausible but unattested entities like a river named
\emph{florida} and a state named \emph{west wyoming}. 

The last example in the ``logical forms'' section is particularly interesting.
The extracted fragment contains \emph{lowest population density} on the natural
language side but only \texttt{density} on the logical form side. However, the
\emph{environment} constrains substitution to happen where appropriate:
this fragment will only be used in cases where the environment already contains
the necessary \texttt{smallest}.

% SEMANTIC PARSING: GEOQUERY / LFS
\begin{table}[t!]
\center
\scalebox{0.96}{
\begin{tabular}{lcc}
\toprule
& Query & Question \\
\midrule
\multicolumn{2}{l}{\textbf{Logical forms}} \\
  seq2seq & 0.62\nosig\nosig \spm{0.07} & 0.76\nosig \spm{0.02} \\
  + Jia et al.\ 16 & 0.61\nosig\nosig \spm{0.03} & {\bf 0.81}\nosig \spm{0.01} \\
  + \thiswork & {\bf 0.65}\sigA\sigB \spm{0.06} & 0.78\sigA \spm{0.01} \\
  + \thiswork{} + concat & 0.63\nosig\nosig \spm{0.04} & 0.79\sigA \spm{0.01} \\
\midrule
\textbf{SQL queries} \\
  Iyer et al.\ 17 \nocite{Iyer17Semparse} & 0.40\nosig\nosig \hspace{25.5pt} &
  0.66\nosig \hspace{25.5pt} \\
  seq2seq & 0.39\nosig\nosig \spm{0.05} & {\bf 0.68}\nosig \spm{0.02} \\
  + \thiswork & {\bf 0.49}\sigA\nosig \spm{0.02} & {\bf 0.68}\nosig \spm{0.02} \\
\bottomrule
\end{tabular}
}
\caption{Meaning representation \emph{exact-match} accuracies on the \geoquery
  dataset. On logical forms, \thiswork approaches the data augmentation approach
  of \citet{Jia16Recombination} on the standard split of the data (``Question'')
  and outperforms it on a split designed to test compositionality (``Query'').
  On SQL expressions, \thiswork leads to substantial improvements on the query
  split and achieves state-of-the-art results. Scores are averaged over 10
  random seeds; the standard deviation across seeds is shown.\textsuperscript{1}
  \sigA Significant improvement over seq2seq baseline ($p < 0.01$).  \sigB
  Significant improvement over \citet{Jia16Recombination} ($p < 0.001$).  (A
  $t$-test is used for LF experiments and a paired binomial test for SQL.)
}
\label{tab:semparse}
\end{table}

Some substitutions are
semantically problematic: for example, the final datapoint in
\autoref{fig:geo-synth} asks about the population of a number (because
substitution has replaced \emph{capital} with \emph{area}); the corresponding
SQL expression would fail to execute. Aside from typing problems, however, the
example is syntactically well-formed and provides correct evidence about
constituent boundaries, alignments and hierarchical structure within the
geography domain. Other synthesized examples (like the second-to-last in
\autoref{fig:geo-synth}) have correct meaning representations but ungrammatical
natural language inputs.

\begin{table}[b!]
  \centering
\begin{tabular}{lcc}
  \toprule
  & Query & Question \\
  \midrule
  \textbf{SQL queries} \\
  seq2seq & 0.03 \spm{0.01} & 0.57 \spm{0.02} \\
  + \thiswork & 0.03 \spm{0.01} & 0.56 \spm{0.02} \\
  \bottomrule
\end{tabular}
  \caption{Negative results: meaning representation accuracies on the
  \textsc{Scholar} dataset. For the query split, synthesized examples do not
  overlap with any of the held-out data; for the question split, they provide
  little information beyond what is already present in the training dataset. In
  both cases a model trained with \thiswork performs indistinguishably from a
  the baseline model.}
  \label{fig:geo-negative}
\end{table}

% LM
\begin{table*}
  \center
\begin{tabular}{lcccccc}
\toprule
& \sc eng & \sc kin & \sc lao & \sc na & \sc pus & \sc tok \\
\midrule
\# train tokens & 2M\nosig & 62K\nosig & 10K\nosig & 28K & 2M\nosig & 30K \\
\midrule
5-MKN & 369\nosig & 241\nosig & 315\nosig & \bf 45.4 & 574\nosig & \bf 44.3 \\
+ \thiswork & \bf 365\sigA & \bf 239\sigA & \bf 313\sigA & \bf 45.4 & \bf 570\sigA & \bf 44.1
\\
\bottomrule
\end{tabular}
\caption{
  Perplexities on low-resource language modeling in English (\textsc{eng}),
  Kinyarwanda (\textsc{kin}), Lao, Na, Pashto (\textsc{pus}) and Tok Pisin
  (\textsc{tok}). Even with a Kneser--Ney smoothed 5-gram model (5-MKN) rather
  than a high-capacity neural model, applying \thiswork leads to small
  improvements in perplexity. \sigA Significant improvement over 5-gram MKN
  baseline (paired binomial test, $p < 0.05$).
}
\label{tab:lm}
\end{table*}

\paragraph{Analysis: dataset statistics}

Applying \thiswork to the \geoquery data increases full example overlap
(described at the end of \autoref{sec:scan}) by 5\% for the question split in
both languages, 6\% for the query split with logical forms, and 9\% for the
query split with SQL expressions, in line with the observation that accuracy
improvements are greater for the query split than the question split.
Augmentation increases token co-occurrence overlap by 3--4\% across all
conditions.

In a larger-scale manual analysis of 100 synthesized examples
from the query split, evaluating them for \textbf{grammaticality} and
\textbf{accuracy} (whether the natural language captures the semantics of the
logical form), we find that 96\% are grammatical, and 98\% are semantically
accurate.

\paragraph{Negative results}

We conclude with a  corresponding set of experiments on the \textsc{Scholar} text-to-SQL
dataset of \citet{Iyer17Semparse}, which is similar to \geoquery in size,
diversity and complexity. In contrast to \geoquery,
however, application of \thiswork to \textsc{Scholar} provides no
improvement. On the query split, there is limited compositional re-use of SQL
sub-queries (in line with the observation of \citet{Finegan18SQL} that average
nesting depth in \textsc{Scholar} is roughly half that of \geoquery).
Correspondingly, full example overlap after augmentation remains at 0\% and
token co-occurrence overlap increases by only 1\%. On the question split, full
example overlap is larger (8\%) but token co-occurrence overlap increases by less
than 1\%. These results suggest that \thiswork is most successful when it can
increase similarity of word co-occurrence statistics in the training and test
sets, and when the input dataset exhibits a high degree of recursion.

\section{Low-resource language modeling}

Both of the previous sections investigated conditional models. The
fragments extracted and reused by \thiswork were essentially synchronous lexicon
entries, in line with example \eref{ex:translation}. We originally motivated
\thiswork with monolingual problems in which we simply wish to improve model
judgments about well-formedness, so we conclude with a set of language modeling
experiments.

We use Wikipedia dumps\footnote[2]{\url{https://dumps.wikimedia.org/}} in five
languages (Kinyarwanda, Lao, Pashto, Tok Pisin, and a subset of English
Wikipedia) as well as the Na dataset of \citet{Adams17LowResLM}. These
languages exhibit the performance of \thiswork across a range of morphological
complexities: for example, Kinyarwanda has a complex noun class system
\cite{Kimenyi80Kinyarwanda} and Pashto has rich derivational morphology
\cite{Tegey1996Pashto}, while Lao and Tok Pisin are comparatively simple
morphologically \cite{Enfield08Lao,Verhaar1995TokPisin}.
Training datasets range from 10K--2M tokens. Like \citeauthor{Adams17LowResLM},
we found that a 5-gram modified Kneser--Ney language model
\cite{Ney94Smoothing} outperformed several varieties of RNN language model, so
we base our \thiswork experiments on the n-gram model instead. We use the
implementation provided in KenLM \cite{Heafield11KenLM}.

We extract fragments with no gaps and a maximum size of 2 tokens, with the
environment taken to be a 2-token window around the extracted fragment. 
New usages are generated only for fragments that occur fewer than 20 times in
the data. In Kinyarwanda, the base dataset contains 3358 sentences. GECA
generates an additional 913, using 913 distinct templates and 199 distinct
fragments.

Rather than training directly on the augmented dataset, as in
preceding sections, we found that the best performance came from training one
language model on the original dataset and one on the augmented dataset, then
interpolating their final probabilities. The weight for this interpolation is
determined on a validation dataset and chosen to be one of $0.05$, $0.1$ and
$0.5$.  

Results are shown in \autoref{tab:lm}. Improvements are not universal
and are more modest than in preceding sections. However, \thiswork decreases
perplexities across multiple languages and never increases them.  These results
suggest that the substitution principle underlying \thiswork is a useful
mechanism for encouraging compositionality even outside conditional tasks
and neural models.

\paragraph{Analysis: examples and statistics}

\begin{figure}[b!]
  \center
  \footnotesize
  \begin{tabularx}{\columnwidth}{>{\RaggedRight\arraybackslash}X}
    \toprule
    \it various copies of portions of the code of hammurabi have been found on baked
    clay tablets , some possibly older than the celebrated basalt stele now in the
    \hlt{night sky} . \\[1em]
    \it the work contains , in an appendix , the german equivalents for the technical
    terms used in the \hlt{glock \$num}~. \\[1em]
    \it \hlt{payments system} in the aclu proposed new directions for the
    organization . \\[1em]

    \it in the \hlt{late triassic} and early nineteenth century , a number of scots-irish
    traders lived among the choctaw and married high-status women . \\
    \bottomrule
  \end{tabularx}
  \caption{Sentences synthesized for the English language modeling task. Most
  examples are syntactically well-formed; some are also semantically plausible.}
  \label{fig:lm-synth}
\end{figure}

In language modeling, \thiswork functions as a smoothing scheme: its primary
effect is to move mass toward n-grams that can appear in productive contexts.
In this sense, \thiswork performs a similar role to the Kneser--Ney smoothing
also used in all LM experiments. With \thiswork, in contrast to Kneser--Ney, the
notion of ``context'' can look forward as well as backward, and capture
longer-range interactions.

Examples of synthesized sentences are shown in \autoref{fig:lm-synth}. Most
sentences are grammatical, and many of the substitutions preserve relevant
semantic type information (substituting locations for locations, times for
times, etc.). However, some ill-formed sentences are also generated.

As in \autoref{sec:semparse}, we manually inspect 100 synthesized sentences.
As before, sentences are evaluated for grammaticality; here, since no explicit
semantics were provided, they are instead evaluated for generic semantic
acceptability. In this case, only 51\% of synthesized sentences are
semantically acceptable, but 79\% are grammatical.

\section{Discussion}

We introduced \thiswork, a simple data augmentation scheme based on identifying
local phrase substitutions that are licensed by common contexts, and demonstrated
that extra training examples generated with \thiswork lead to substantial
improvements on both diagnostic and natural datasets for semantic parsing and
language modeling. 

While the approach is surprisingly effective in its current
form, we view these results primarily as an invitation to consider more
carefully the role played by representations of sentence fragments in larger
questions about compositionality in black-box sequence models. The procedure
detailed in this paper relies on exact string matching to identify common
context; future work might take advantage of learned representations of spans
and their environments \citep{Mikolov13Embeddings,Peters18ELMO}. Further
improvements are likely obtainable by constraining the extracted fragments to
respect constituent boundaries when syntactic information is available.

The experiments presented here focus on rewriting sentences using evidence
\emph{within} a dataset to encourage generalization to new outputs. An
alternative line of work on paraphrase-based data augmentation
\cite{Ganitkevitch13PPDB,Iyyer18SynParaphrase} uses external, text-only
resources to encourage robust interpretation of new inputs corresponding to
known outputs.  The two lines of work could be combined, e.g.\ by using
\thiswork-identified fragments to indicate productive locations for
sub-sentential paraphrasing.

More generally, the present results underline the extent to which current models
fail to learn simple, context-independent notions of reuse, but also how easy it
is to make progress towards addressing this problem without fundamental changes
in model architecture.

\section*{Reproducibility}

Code for all experiments in this paper may be found at
\url{github.com/jacobandreas/geca}. 

\section*{Acknowledgments}
Thanks to Oliver Adams for assistance with the language modeling experiments,
and to the anonymous reviewers for suggestions in the analysis sections.

%\newpage

\bibliographystyle{acl-natbib}
\bibliography{jacob}

\end{document}